\documentclass[review]{elsarticle}

\usepackage{hyperref}
\usepackage{graphicx}
\usepackage{amsmath}
\usepackage{amssymb}
\usepackage{geometry}
\usepackage{longtable}
\usepackage{subcaption}
\usepackage[english]{babel}
\usepackage{algorithm}
\usepackage{algpseudocode}
\usepackage[usenames,dvipsnames,svgnames,table]{xcolor}

\algnewcommand{\Initialize}[1]{%
  \State \textbf{Initialize:}
  \Statex \hspace*{\algorithmicindent}\parbox[t]{.8\linewidth}{\raggedright #1}
}
\makeatletter
\algnewcommand{\Stepb}[1]{\Statex \hskip\ALG@thistlm \(\triangleright\) #1}
\makeatother
\algnewcommand{\Stepa}[1]{\State #1}

\newif\ifnote
\notetrue

\newcommand{\nada}[1]{\textcolor{black}{{{}#1}}}
\newcommand{\nadb}[1]{\textcolor{black}{{{}#1}}}

\newtheorem{theorem}{Theorem}
\newtheorem{lemma}[theorem]{Lemma}

\journal{Journal of \LaTeX\ Templates}

\begin{document}

\begin{frontmatter}

\title{Distributed learning of deep neural network over multiple agents}

\author[add1]{Otkrist Gupta\corref{mycorrespondingauthor}}
\cortext[mycorrespondingauthor]{Corresponding author}
\ead{otkrist@mit.edu}
\author[add1]{Ramesh Raskar}

\address[add1]{Massachusetts Institute of Technology\\ 77 Massachusetts Ave, Cambridge MA 02139, USA}

\begin{abstract}
In domains such as health care and finance, shortage of labeled data and computational resources is a critical issue while developing machine learning algorithms. To address the issue of labeled data scarcity in training and deployment of neural network-based systems, we propose a new technique to train deep neural networks over several data sources. Our method allows for deep neural networks to be trained using data from multiple entities in a distributed fashion. We evaluate our algorithm on existing datasets and show that it obtains performance which is similar to a regular neural network trained on a single machine. We further extend it to incorporate semi-supervised learning when training with few labeled samples, and analyze any security concerns that may arise. Our algorithm paves the way for distributed training of deep neural networks in data sensitive applications when raw data may not be shared directly.
\end{abstract}


\end{frontmatter}


\section{Introduction}\label{sec:introduction}

Deep neural networks have become the new state of the art in classification and prediction of high dimensional data such as images, videos and bio-sensors. Emerging technologies in domains such as biomedicine and health stand to benefit from building deep neural networks for prediction and inference by automating the human involvement and reducing the cost of operation. However, training of deep neural nets can be extremely data intensive requiring preparation of large scale datasets collected from multiple entities \cite{chervenak2000data,chuang2000distributed}. A deep neural network typically contains millions of parameters and requires tremendous computing power for training, making it difficult for individual data repositories to train them. 

Sufficiently deep neural architectures needing large supercomputing resources and engineering oversight may be required for optimal accuracy in real world applications. Furthermore, application of deep learning to such domains can sometimes be challenging because of privacy and ethical issues associated with sharing of de-anonymized data. While a lot of such data entities have vested interest in developing new deep learning algorithms, they might also be obligated to keep their user data private, making it even more challenging to use this data while building machine learning pipelines. In this paper, we attempt to solve these problems by proposing methods that enable training of neural networks using multiple data sources and a single supercomputing resource.

\section{Related Work}

Deep neural networks have proven to be an effective tool to classify and segment high dimensional data such as images\cite{krizhevsky2012imagenet}, audio and videos\cite{karpathy2015deep}. Deep models can be several hundreds of layers deep\cite{he2016deep}, and can have millions of parameters requiring large amounts of computational resources, creating the need for research in distributed training methodologies\cite{dean2012large}. Interesting techniques include distributed gradient optimization\cite{mcdonald2009efficient,zinkevich2010parallelized}, online learning with delayed updates\cite{langford2009slow} and hashing and simplification of kernels\cite{shi2009hash}. Such techniques can be utilized to train very large scale deep neural networks spanning several machines\cite{agarwal2011distributed} or to efficiently utilize several GPUs on a single machine\cite{agarwal2014reliable}. In this paper we propose a technique for distributed computing combining data from several different sources.

Secure computation continues to be a challenging problem in computer science \cite{sood2012combined}. One category of solutions to this problem involve adopting oblivious transfer protocols to perform secure dot product over multiple entities in polynomial time \cite{avidan2006blind}. While this method is secure, it is somewhat impractical when considering large scale datasets because of resource requirements. A more practical approach proposed in \cite{avidan2006blind} involves sharing only SIFT and HOG features instead of the actual raw data. However, as shown in \cite{dosovitskiy2015inverting}, such feature vectors can be inverted very accurately using prior knowledge of the methods used to create them. Neural networks have been shown to be extremely robust to addition of noise and their denoising and reconstruction properties make it difficult to compute them securely \cite{vincent2010stacked}. Neural networks have also been shown to be able to recover an entire image from only a partial input \cite{pathak2016context}, rendering simple obfuscation methods inert.

Widespread application of neural networks in sensitive areas such as finance and health, has created a need to develop methods for both distributed and secure training \cite{secretan2007privacy,chonka2011cloud,wu2007secure} and classification in neural networks. Under distributed and secure processing paradigms, the owner of the neural network doesn't have access to the actual raw data used to train the neural network \cite{barni2006privacy}. \nada{This also includes secure paradigms in cloud computing \cite{karam2012security,subashini2011survey}, virtualization \cite{mackay2012security} and service oriented architectures \cite{baker2015security}}. The secure paradigms may also extend to the neural activations and (hyper)parameters. Such algorithms form a subset inside the broader realm of multi-party protocol problems involving secure computation over several parties \cite{goldreich1987play, yao1986generate}. Some interesting solutions include using Ada-boost to jointly train classifier ensembles \cite{zhang2013privacy}, using random rotation perturbations for homomorphic pseudo-encryption \cite{chen2005random} and applying homomorphic cryptosystem to perform secure computation \cite{orlandi2007oblivious}.


\begin{figure*}[t]
\begin{subfigure} {0.5\textwidth}
\includegraphics[width=1\textwidth,trim={0cm 0 0cm 0},clip]{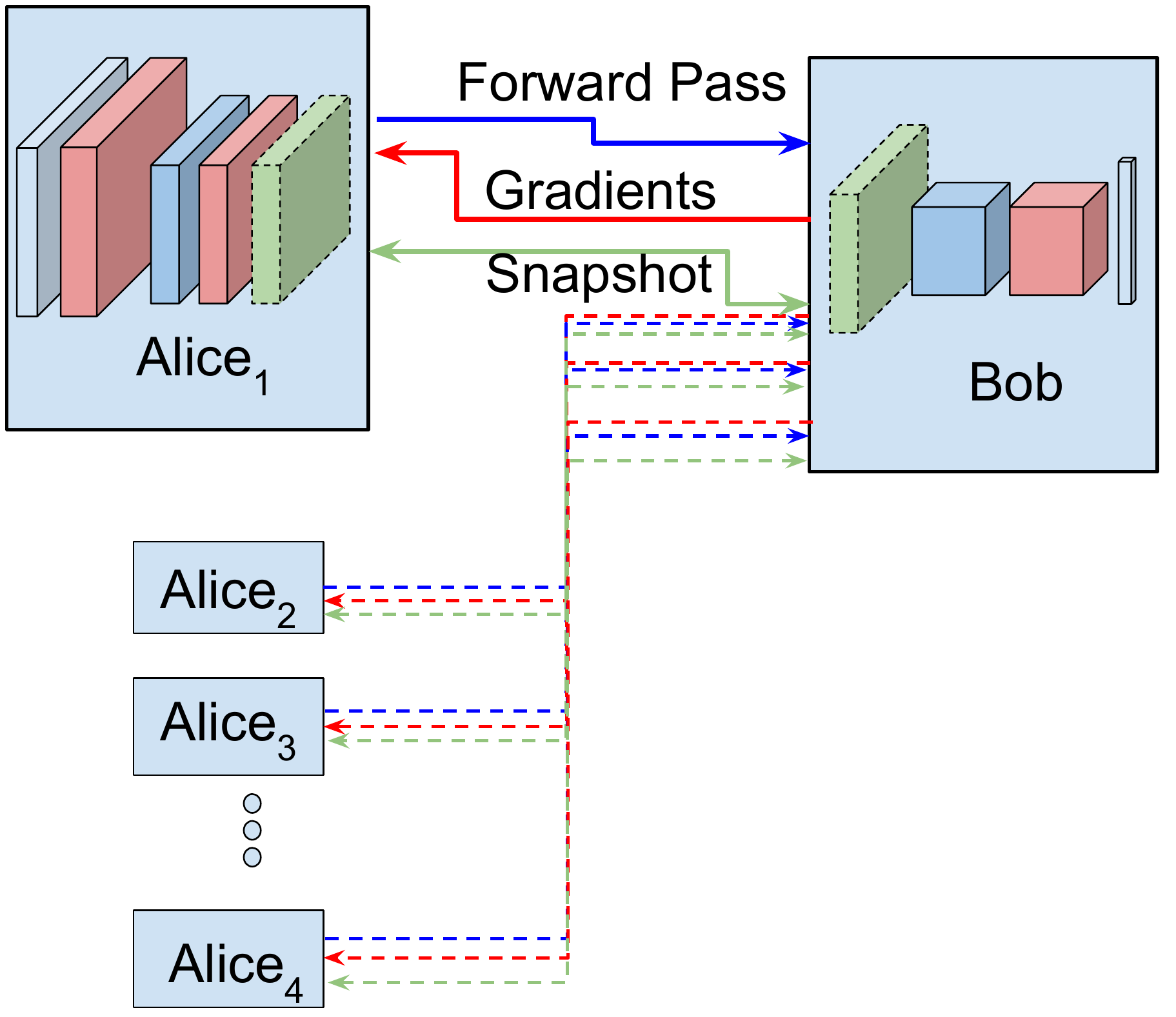}
\caption{Centralized distributed neural network training.}
\label{centralized}
\end{subfigure}
\begin{subfigure} {0.5\textwidth}
\includegraphics[width=1\textwidth,trim={0cm 0 0cm 0},clip]{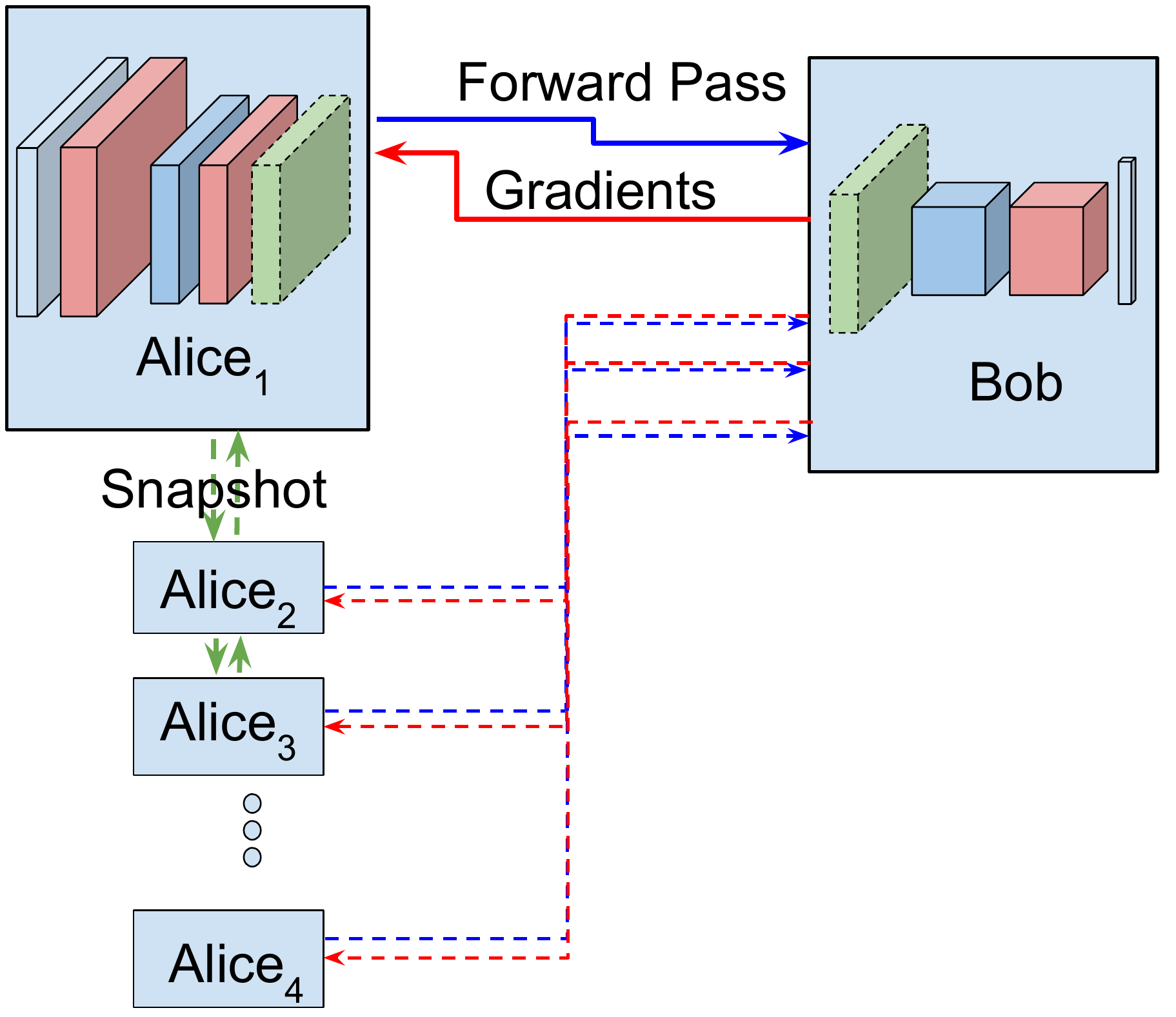}
\caption{Peer-to-peer training for distributed learning.}
\label{p2p}
\end{subfigure}
\caption{Two modalities of our algorithm: centralized mode (\ref{centralized}) and peer-to-peer mode (\ref{p2p}).}
\label{wireframe}
\end{figure*}

\section{Theory}

\nadb{In this paper we propose new techniques that can be used to train deep neural networks over multiple data sources while mitigating the need to share raw labeled data directly. Specifically we address the problem of training a deep neural network over several data entities (Alice(s)) and one supercomputing resource (Bob). We aim at solving this problem while satisfying the following requirements :}

\begin{enumerate}
   \item A single data entity (Alice) doesn't need to share the data with Bob or other data resources. 
   \item The supercomputing resource (Bob) wants control over the architecture of the Neural Network(s)
   \item Bob also keeps a part of network parameters required for inference.
\end{enumerate}

\nadb{In upcoming sections we will show how to train neural networks between multiple data entities (Alice(s)) and a supercomputing resource (Bob). Techniques will include methods which encode data into a different space and transmit it to train a deep neural network. We will further explore how a third-party can use this neural network to classify and perform inference. Our algorithm can be run using one or multiple data entities, and can be run in peer-to-peer or centralized mode. Please see Figure \ref{wireframe} for the schematic depiction of algorithm modalities.}


\subsection{Distributed training over single entity}
\label{sec:single}

\begin{algorithm}[h]
\small
\caption{Distributed Neural Network training over 2 agents.}
\label{alg1}
\begin{algorithmic}[1]
\Initialize{$\phi \gets$ Random Initializer (Xavier/Gaussian) \\
         $F_a \gets \{L_0, L_1, ... L_n\}$ \\
            $F_b \gets \{L_{n+1}, L_{n+2}, ...L_N\}$ \\
            }
\State Alice randomly initializes the weights of $F_a$ using $\phi$
\State Bob randomly initializes the weights of $F_b$ using $\phi$
\While{Alice has new data to train on}
\Stepa{Alice uses standard forward propagation on data}
\Stepb{$X \gets F_a(data)$}
\Stepa{Alice sends $n^{th}$ layer output $X$ and label to Bob}
\Stepb{$Send((X, label),Bob)$.}
\Stepa{Bob propagates incoming features on its network}
\Stepb{$output \gets F_b(X)$}
\Stepa{Bob generates gradients for its final layer}
\Stepb{$gradient \gets G'(output, label)$}
\Stepa{Bob backpropagates the error in $F_b$ until $L_{n+1}$}
\Stepb{$F_b', gradient' \gets F_b^T(gradient)$}
\Stepa{Bob sends gradient of $L_n$ to Alice}
\Stepb{$Send(gradient', Alice)$}
\Stepa{Alice backpropagates gradients received}
\Stepb{$F_a',\_ \gets F_a^T(gradient')$}
\EndWhile
\end{algorithmic}
\end{algorithm}

We will start by describing the algorithm in its simplest form which considers training a neural network using data from a single entity and supercomputing resource. Let us define a deep neural network as a function $F$, topologically describable using a sequence of layers $\{L_0, L_1, ... L_N\}$. For a given input ($data$), the output of this function is given by $F(data)$ which is computed by sequential application of layers $F(data) \gets L_N(L_{N-1}...(L_0(data)))$.

Let $G_{loss}(output, label)$ denote the customized loss function used for computing gradients for the final layer. Gradients can be backpropagated over each layer to generate gradients of previous layers and to update the current layer. We will use $L_{i}^T(gradient)$ to denote the process of backpropagation over one layer and $F^T(gradient)$ to denote backpropagation over the entire Neural Network. Similar to forward propagation, backpropagation on the entire neural network is comprised of sequential backward passes $F^T(gradient) \gets L_1^T(L_{2}^T...(L_N^T(gradient)))$. Please note that the backward passes will require activations after the forward pass on individual perceptrons.

Finally, $Send(X, Y)$ represents the process of sending data $X$ over the network to entity $Y$. In the beginning, Alice and Bob initialize their parameters randomly. Alice then iterates over its dataset and transmits encoded representations to Bob. Bob then computes losses and gradients and sends the gradients back to Alice. Algorithm \ref{alg1} describes how to train a deep neural classifier using a single data source.

\subsubsection{Correctness}
\label{correctnessone}

Here we analyze if training using our distributed algorithm produces the same results as a normal training procedure. Under a normal training procedure we would first compute forward pass $output \gets F(data)$ followed by computation of loss gradients $gradients \gets G(output, label)$. These gradients will be backpropagated to refresh weights $F' \gets F^T(gradients)$.

Since forward propagation involves sequential application of individual layers we concur that $F(data)$ is same as $F_b(F_a(data))$. Therefore the process of sequential computation and transmission followed by computation of remaining layers is functionally identical to application of all layers at once. Similarly because of the chain rule in differentiation, backpropagating $F^T(gradients)$ is functionally identical to sequential application of $F_a^T(F_b^T(gradients))$. Therefore, we can conclude that our algorithm will produce identical results to a normal training procedure.

\begin{algorithm}[h!]
\small
\caption{Distributed Neural Network over N+1 agents.}
\label{alg2}
\begin{algorithmic}[1]
\Initialize{$\phi \gets$ Random Initializer (Xavier/Gaussian) \\
      $F_{a,1} \gets \{L_0, L_1, ... L_n\}$ \\
            $F_b \gets \{L_{n+1}, L_{n+2}, ...L_N\}$ \\
            }
\State Alice$_{1}$ randomly initializes the weights of $F_{a,1}$ using $\phi$
\State Bob randomly initializes the weights of $F_b$ using $\phi$
\State Bob sets Alice$_1$ as last trained
\While{Bob waits for next Alice$_{j}$ to send data}
\Stepa{Alice$_{j}$ requests Bob for last Alice$_o$ that trained}
\Stepa{Alice$_{j}$ updates its weights}
\Stepb{$F_{a,j} \gets F_{a,o}$}
\Stepa{Alice$_{j}$ uses standard forward propagation on data}
\Stepb{$X \gets F_{a,j}(data)$}
\Stepa{Alice$_j$ sends $n^{th}$ layer output and label to Bob}
\Stepb{$Send((X, label),Bob)$.}
\Stepa{Bob propagates incoming features on its network}
\Stepb{$output \gets F_b(X)$}
\Stepa{Bob generates gradients for its final layer}
\Stepb{$gradient \gets G'(output, label)$}
\Stepa{Bob backpropagates the error in $F_b$ until $L_{n+1}$}
\Stepb{$F_b', gradient' \gets F_b^T(gradient)$}
\Stepa{Bob sends gradient of $L_n$ to Alice$_j$}
\Stepb{$Send(gradient', Alice_j)$}
\Stepa{Alice$_j$ backpropagates the gradients it received}
\Stepb{$F_{a,j}',\_ \gets F_{a,j}^T(gradient')$}
\Stepa{Bob sets Alice$_j$ as last trained}
\EndWhile
\end{algorithmic}
\end{algorithm}

\subsection{Distributed training over multiple entities}
Here we demonstrate how to extend the algorithm described in \ref{sec:single} to train using multiple data entities. We will use the same mathematical notations as used in \ref{sec:single} when defining neural network forward and backward propagation. In algorithm \ref{alg2} we demonstrate how to extend our algorithm when there are $N$ data entities, each of them is denoted by $Alice_i$. 

In algorithm \ref{alg2} at the first initialization step, Bob sends Alice$_1$ topological description of first $N$ layers. Alice and Bob use standard system level libraries for random initialization of their parameters. Bob then sets Alice$_1$ as the last agent used for training and begins training using data from Alice$_1$. We modify \ref{alg1} and add a step which uses data from multiple entities in a round robin fashion, allowing for a distributed learning framework. However, for consistency, Alice$_j$ may be required to update weights before they begin their training. We solve this by providing two separate methodologies involving peer-to-peer and centralized configurations. \nada{In the \textit{centralized} mode, Alice uploads an encrypted weights file to either Bob or a third-party server. When a new Alice wishes to train, it downloads and decrypts these weights. In \textit{peer-to-peer} mode, Bob sends the last trained Alice's address to the current training party and Alice uses this to connect and download the encrypted weights. The implementation details for both methods can be seen in supplementary material.} Once the weights are updated, Alice$_j$ continues its training. \nada{Since the same weights are initialized in both centralized and peer-to-peer mode, the final result of training is identical in both modalities.}

\subsubsection{Correctness}
\label{sec:correctness}

We analyze if training using our algorithm produces results which are identical when training with all the data combined on a single machine (under the assumption that the data arriving at multiple entities preserves the order and random weights use same initialization). The algorithm correctness stems from the fact that Bob and at least one of Alice$_o$ have identical neural network parameters to regular training at iteration$_k$. We use inductive techniques to prove that this is indeed the case.

\begin{lemma}
The neural network being trained at iteration$_k$ is identical to the neural network if it was trained by just one entity.
\label{lemma:correctness}
\end{lemma}

\textbf{Base Case:} One of Alice$_{1...N}$ has the correct weights at beginning of first iteration.

\textbf{Proof:} Alice$_1$ randomly initialized weights and Bob used these weights during first iteration. We assume that this initialization is consistent when training with single entity. In case another Alice$_{j}$ attempts to train, it will refresh the weights to correct value.

\textbf{Recursive Case:}
Assertion: If Alice$_{j}$ has correct weights at beginning of iteration$_i$ it will have correct weights at beginning of iteration $i+1$. 

\textbf{Proof:} Alice$_{j}$ performs backpropagation as the final step in iteration $i$. Since this backpropagation is functionally equivalent to backpropagation applied over the entire neural network at once, Alice$_{j}$ continues to have correct parameters at the end of one training iteration. ($F^T(gradient)$ is functionally identical to sequential application of $F_{a,j}^T(F_b^T(data))$, as discussed in \ref{correctnessone}).

\subsection{Semi-supervised application}
In this section we describe how to modify the distributed neural network algorithm to incorporate semi-supervised learning and generative losses when training with fewer data points. In situations with fewer labeled data-samples, a reasonable approach includes learning hierarchical representations using unsupervised learning \cite{shin2013stacked}. Compressed representations generated using unsupervised learning and \textit{autoencoders} can be used directly for classification \cite{coates2012emergence}. Additionally, we can combine the losses of generative and predictive segments to perform semi supervised learning, adding a regularization component while training on fewer samples \cite{weston2012deep}.

\begin{algorithm}[h!]
\small
\caption{Distributed Neural Network with an Autoencoder over N+1 agents.}
\label{alg3}
\begin{algorithmic}[1]
\Initialize{$\phi \gets$ Random Initializer (Xavier/Gaussian) \\
      $F_{e,1} \gets \{L_0, L_1, ... L_m\}$ \\
      $F_{d,1} \gets \{L_m, L_{m+1}, ... L_n\}$ \\
            $F_b \gets \{L_{n+1}, L_{n+2}, ...\}$ \\
            }
\State Alice$_{1}$ randomly initializes the weights of $F_{a,1}$ using $\phi$
\State Bob randomly initializes the weights of $F_b$ using $\phi$
\State Alice$_{1}$ transmits weights of $F_{a,1}$ to Alice$_{2...N}$
\While{Bob waits for next feature vector from Alice$_{j}$}
\Stepa{Alice$_{j}$ requests Bob for last Alice$_o$ that trained}
\Stepa{Alice$_{j}$ updates its weights}
\Stepb{$F_{a,j} \gets F_{a,o}$}
\Stepa{Alice$_{j}$ uses standard forward propagation on data}
\Stepb{$X_m \gets F_{e,j}(data)$}
\Stepb{$X \gets F_{d,j}(X_m)$}
\Stepa{Alice$_{j}$ sends $m^{th}$ layer output and label to Bob}
\Stepb{$Send((X_m, label), Bob)$.}
\Stepa{Bob propagates incoming features on its network $F_{b}$}
\Stepb{$output \gets F_b(X_m)$.}
\Stepa{Bob generates gradient for its final layer}
\Stepb{$gradient \gets G'(output, label)$}
\Stepa{Bob backpropagates the error in $F_b$ until $L_{n+1}$}
\Stepb{$F_b', gradient' \gets F_b^T(gradient)$}
\Stepa{Bob sends gradient for $L_{n}$ to Alice$_j$}
\Stepb{$Send(gradient', Alice_j)$}
\Stepa{Alice$_{j}$ generates autoencoder gradient for its decoder}
\Stepb{$F_{d,j}', gradient'_{enc} = F_{d,j}^T(X)$}
\Stepa{Alice$_{j}$ backpropagates combined gradients}
\Stepb{$F_a,\_ \gets F_a^T(\eta(gradient', gradient'_{enc}))$}
\Stepa{Bob sets Alice$_j$ as last trained}
\EndWhile
\end{algorithmic}
\end{algorithm}

\nada{Over here we demonstrate how we can train autoencoders and semi-supervised learners using a modified version of algorithm \ref{alg1}. Such unsupervised learning methods can be extremely helpful when training with small amounts of labeled data.} We assume that out of $n$ layers for Alice, the first $m$ layers are encoder and the remaining $n-m$ layers belong to its decoder. $F_{e,i}$ denotes the forward propagation over encoder (computed by sequential application $L_m(L_{m-1}...(L_0(data)))$). $F_{d,i}$ denotes application of decoder layers. During forward propagation Alice propagates data through all $n$ layers and sends output from $m^{th}$ layer to Bob. Bob propagates the output tensor from Alice through $L_{n...N}$ and computes the classifier loss (logistic regression).

Let $loss$ define the logistic regression loss in the predictive segment of the neural network (last $N-n$ layers owned by Bob), and let $loss_{enc}$ define the contrastive loss in autoencoder (completely owned by Alice(s)). Bob can compute $loss$ using its softmax layer and can back-propagate gradients computed using this loss to layer $L_{n+1}$ giving gradients from classifier network [$gradient' \gets F_b^T(gradient)$]. Alice$_i$ can compute the autoencoder gradients and can backpropagate it through its \textit{decoder} network [$F_{d,i}^T(gradient_{enc})$]. We can facilitate semi-supervised learning by combining a weighted sum of two losses. The weight $\alpha$ is an added hyperparameter which can be tuned during training.

\begin{equation}
  \eta \gets F_b^T(gradient) + \alpha*F_{d,i}^T(gradient_{enc})
\end{equation}

After the initialization steps, Alice propagates its data through its network and sends output from the encoder part to Bob. Bob does a complete forward and backward to send gradients to Alice. Alice then combines losses from its decoder network with gradients received from Bob and uses them to perform backpropagation (please see algorithm \ref{alg3} for detailed description).

\subsection{Online learning}
An additional advantage of using our algorithm is that the training can be performed in an online fashion by providing Bob output of forward propagation whenever there is new annotated data. In the beginning instead of transmitting the entire neural net, Alice$_i$ can initialize the weights randomly using a seed and just send the seed to Alice$_{1...N}$ preventing further network overhead. When Alice is requested for weights in peer-to-peer mode, it can simply share the \textit{weight updates}, which it adds to its parameters during the course of training. The combined value of weight updates can be computed by subtracting weights at beginning of training from current weights. For security, Alice can also upload the encrypted weight updates to a centralized weight server, making it harder to reverse engineer actual weights when using man-in-middle attack. Weights can be refreshed by Alice by combining its initial weights with subsequent weight updates downloaded from the centralized weight server (or Alice(s) depending on mode).  To facilitate centralized modality, we can modify step 6 of algorithm \ref{alg2}, replacing it with a request to download encrypted weights from weight server. Once training is over Alice$_j$ can upload the new encrypted weights to the weight server (please refer to step 15 in algorithm \ref{alg2}).

\subsection{Analyzing Security Concerns}
While a rigorous information theoretical analysis of security is beyond the scope of this paper, over here we sketch out a simple explanation of why reconstructing the data sent by Alice is extremely challenging. The algorithm security lies in whether Bob can invert parameters ($F_{a}$) used by Alice during the forward propagation. Bob can indeed build a decoder for compressed representations transmitted by Alice, but it requires Alice revealing the current parameters of its section of neural network \cite{dosovitskiy2015inverting}.

In this section we make an argument that Bob cannot discover the parameters used by Alice as long as its layers (denoted by $F_{a}$) contain at least one fully connected layer. We will use the word ``\textit{configuration}'' to denote an isomorphic change in network topology which leads to functionally identical neural network.

\begin{lemma}
\label{lemma:sec1}
Let layer M be a fully connected layer containing N outputs then layer M has at least N! functionally equivalent ``\textit{configurations}''.
\end{lemma}
\textbf{Proof:} We construct a layer M and transpose N output neurons. The output of neurons is reordered without changing weights or affecting learning in any way. Since there are $N!$ possible orderings of these neurons at least $N!$ unique \textit{configurations} are possible depending on how the weights were initialized.

Bob will have to go through at least $N!$ possible \textit{configurations} to invert the transformation applied by Alice. Since $N! > (N/2)^N >e^N$ this will require an exponential amount of time in a layer of size $N$. For example if the fully connected layer has 4096 neurons and each configuration could be tested in a second, it would take Bob more than the current age of the universe to figure out parameters used by Alice.

\begin{figure*}[t]
\begin{subfigure} {0.5\textwidth}
\includegraphics[width=\textwidth]{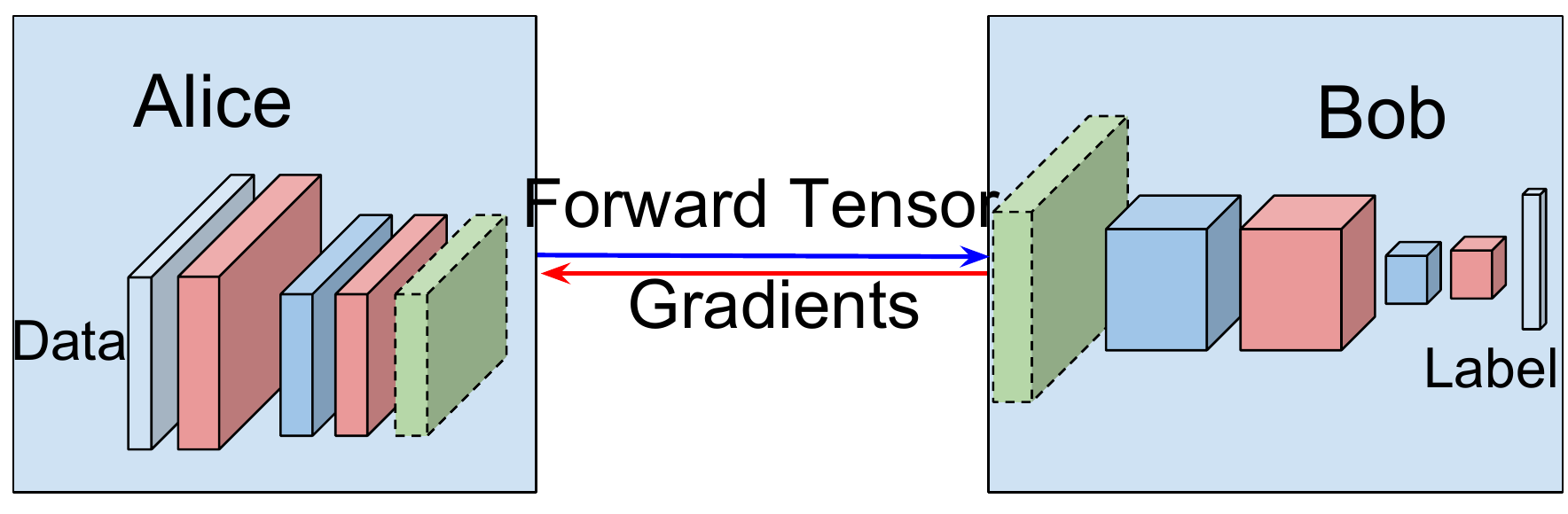}
\caption{Training with label propagation}
\label{wlabel}
\end{subfigure}
\begin{subfigure} {0.5\textwidth}
\includegraphics[width=\textwidth]{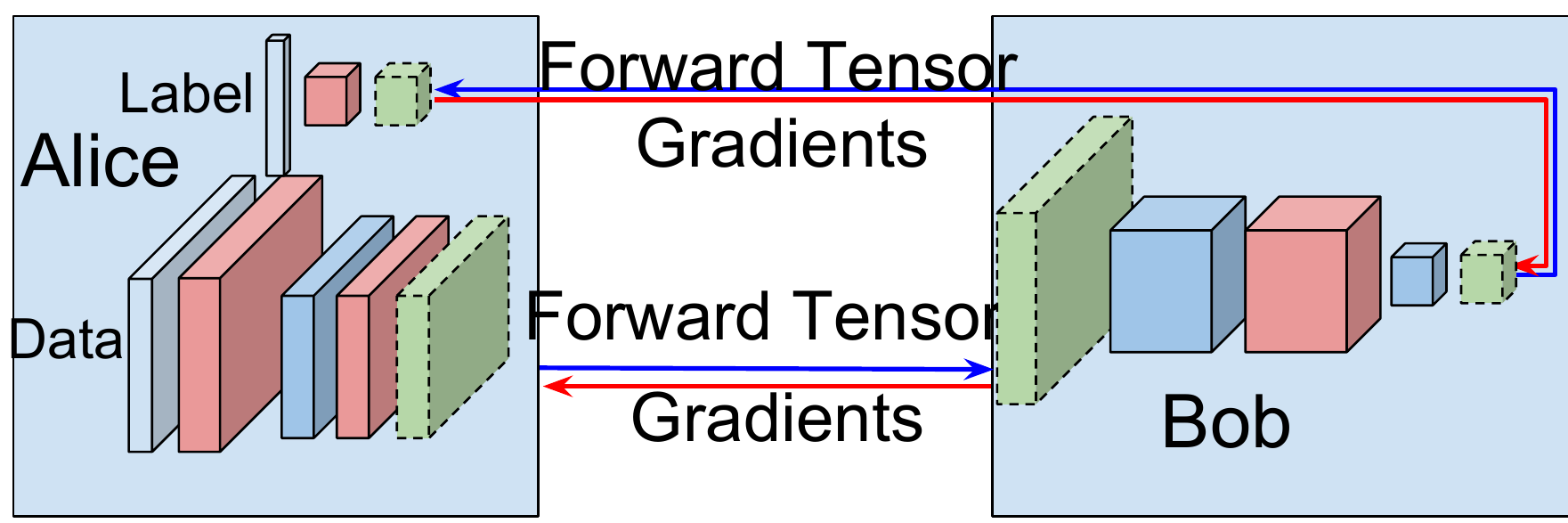}
\caption{Training without label sharing}
\label{wolabel}
\end{subfigure}
\caption{Figure (\ref{wlabel}) shows the normal training procedure while figure (\ref{wolabel}) demonstrates how to train without transmitting labels, by wrapping the network around at its last layers.}
\label{withoutlabel}
\end{figure*}

\subsection{Training without label propagation}

While the algorithm we just described doesn't require sharing raw data, it still does involve sharing labels. We can mitigate this problem by presenting a simple adjustment to the training framework. In this topological modification, we wrap the network around at its end layers and send those back to Alice (see figure \ref{withoutlabel}). While Bob still retains majority of its layers, it lets Alice generate the gradients from the end layers and uses them for backpropagation over its own network. We can use a similar argument as one used in lemma \ref{lemma:correctness} to prove that this method will still work after the layers have been wrapped around. Please see figure \ref{withoutlabel} for a schematic description of our training methodology without label sharing.

\section{Datasets and Implementation}

We use standard \textit{json} communication libraries for asynchronous RPC for implementation. On top of those, we implement a custom protocol for training once a secure connection is established using SSL. Our protocol defines several network primitives (implemented as remote functions) which we broadly divide in 3 parts (1) Training request, (2) Tensor transmission and (3) Weight update. Please refer to appendix for a complete list of network primitives. We describe these three network primitives categories in our supplementary material.

\subsection{Mixed NIST}
Mixed NIST (MNIST) database \cite{lecun1989backpropagation} contains handwritten digits sampled from postal codes and is a subset of a much larger dataset available from the National Institute Science and Technology. MNIST comprises of a total of 70,000 samples divided into 60,000 training samples and 10,000 testing samples. Original binary images were reformatted and spatially normalized to fit in a $20 \times 20$ bounding box. Anti-aliasing techniques were used to convert black and white (bilevel) images to grey scale images. Finally the digits were placed in a $28 \times 28$ grid, by computing the center of mass of the pixels and shifting and superimposing images in the center of a $28 \times 28$ image.

\subsection{Canadian Institute For Advanced Research}
The Canadian Institute For Advanced Research (CIFAR-10) dataset is a labeled subset of tiny images dataset (containing 80 million images). It is composed of 60,000, $32 \times 32$ color images distributed over 10 different class labels. The dataset consists of 50,000 training samples and 10,000 testing images. Images are uniformly distributed over 10 classes with training batches containing exactly 6000 images for each class. The classes are mutually exclusive and there are no semantic overlaps between the images coming from different labels. We normalized the images using GCA whitening and applied global mean subtraction before training. The same dataset also includes a 100 class variation referred to as CIFAR-100.

\subsection{ILSVRC (ImageNet) 2012}
\nada{This dataset includes approximately 1.2 million images labeled with the presence or absence of 1000 object categories. It also includes 150,000 images for validation and testing purposes. The 1000 object categories are a subset of a larger dataset (ImageNet), which includes 10 million images spanning 10,000 object categories. The object categories may be internal or leaf nodes but do not overlap. The dataset comprises images with varying sizes which are resized to $256 \times 256$ and mean subtracted before training.}

\section{Experiments and Applications}

We implement our algorithm and protocol using python bindings for \textit{caffe}\cite{jia2014caffe}. We test our implementation on datasets of various sizes (50K - 1M) and classes (10, 100 or 1000 classes). We demonstrate that our method works across a range of different topologies and experimentally verify identical results when training over multiple agents. \nada{All datasets were trained for an equal number of epochs for fair evaluation.}

In \ref{sec:correctness} we show why our algorithm should give results identical to a normal training procedure. We experimentally verify our method's correctness by implementing it and training it on a wide array of datasets and topologies including MNIST, ILSVRC 12 and CIFAR 10. Table \ref{accuracy_table} lists datasets and topologies combined with their test accuracies. \nada{Test accuracies are computed by comparing the number of correctly labeled samples to the total number of test data points.}. As shown in table \ref{accuracy_table}, the network converges to similar accuracies when training over several agents in a distributed fashion.

\begin{table*}[t]
\centering
  \begin{tabular}{ l | l | l | l | l }
    Dataset & Topology & Accuracy (Single Agent) & Accuracy using our method & Epochs \\ \hline
    MNIST & LeNet \cite{lecun1989backpropagation}& 99.18 \% & 99.20 \% & 50 \\
    CIFAR 10 & VGG \cite{simonyan2014very} & 92.45 \% & 92.43 \% & 200 \\
    CIFAR 100 & VGG \cite{simonyan2014very} & 66.47 \% & 66.59 \% & 200 \\
    ILSVRC 12 & AlexNet \cite{krizhevsky2012imagenet} & 57.1 \% & 57.1 \% & 100 \\
  \end{tabular}
\caption{Accuracies when training using multi-agent algorithm vs when training on a single machine.}
\label{accuracy_table}
\end{table*}


\subsection{Comparison with existing methods}

\begin{figure*}[t]
\begin{subfigure} {0.5\textwidth}
\includegraphics[width=1\textwidth,trim={0.0cm 0 0.0cm 0},clip]{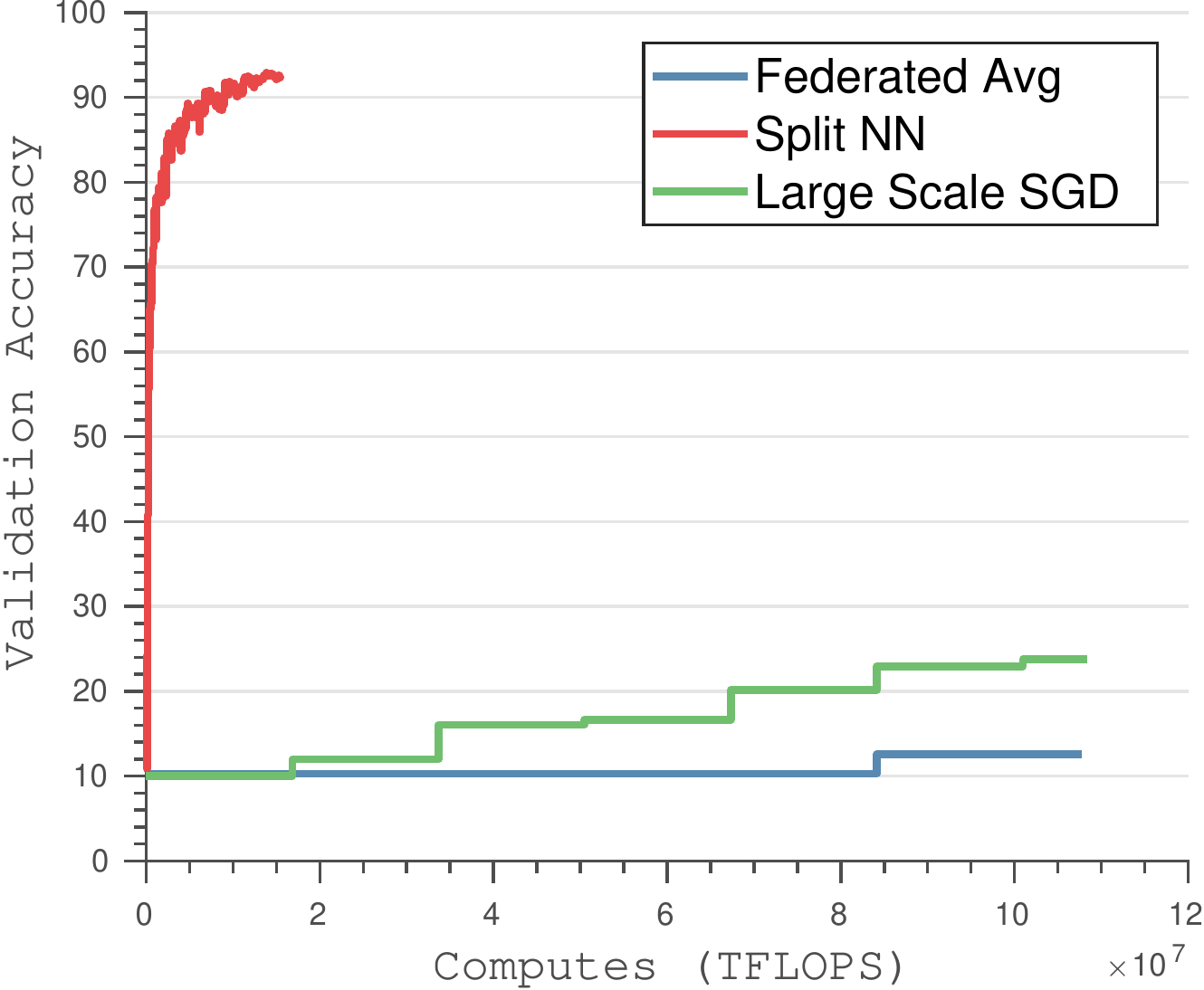}
\caption{Validation accuracy with client side flops when training 100 clients (VGG and CIFAR 10).}
\label{cifar100computeb}
\end{subfigure}
\begin{subfigure} {0.5\textwidth}
\includegraphics[width=1\textwidth,trim={0.0cm 0 0.0cm 0},clip]{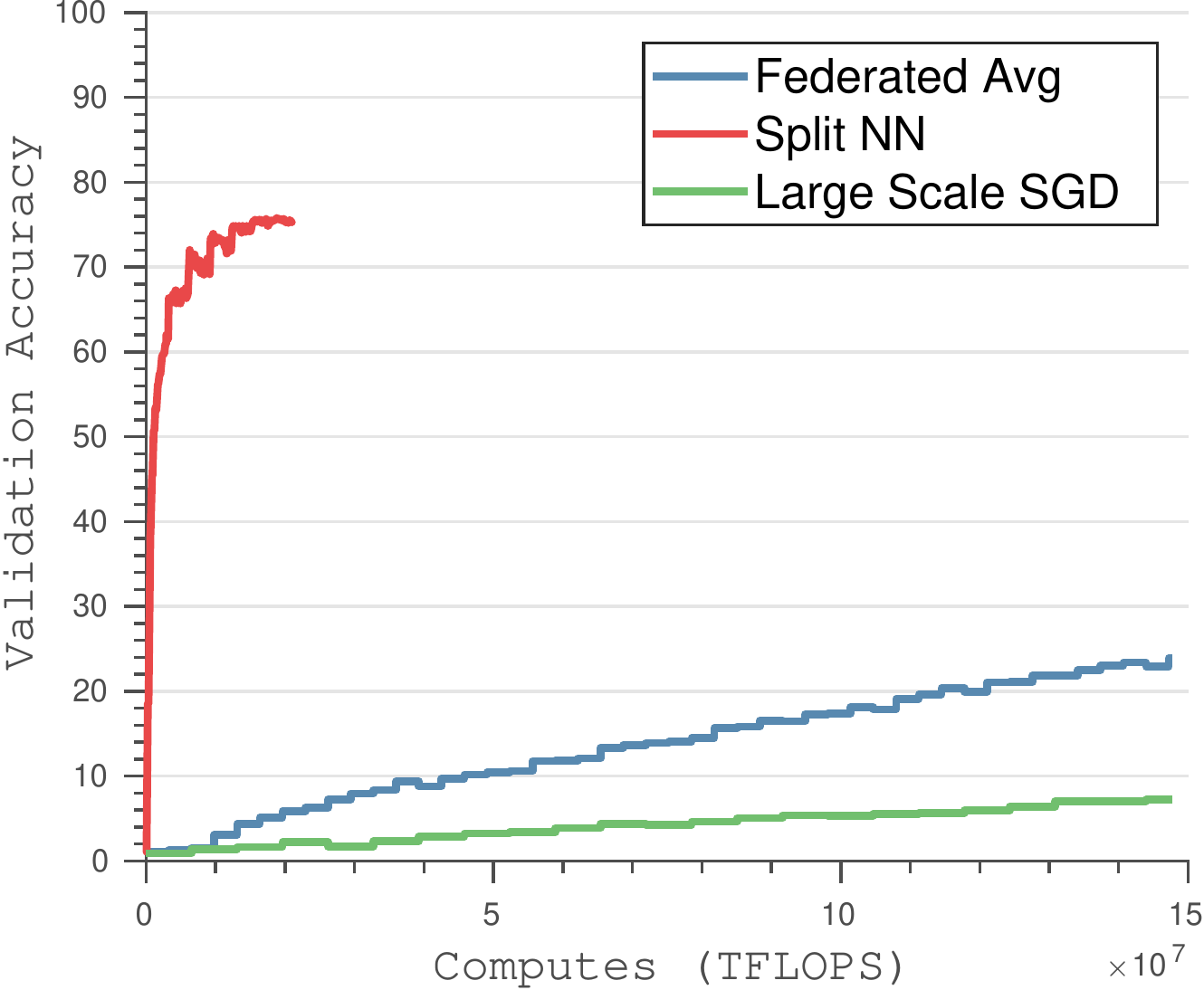}
\caption{Validation accuracy with client side flops when training 500 clients (Resnet-50 and CIFAR 100).}
\label{cifarlimitation10client}
\end{subfigure}
\caption{Comparison of client side computational cost of our method against existing state of the art methods.}
\label{cifar100compute}
\end{figure*}

\nada{We compare our method against the modern state-of-the-art methods including large-batch global SGD \cite{chen2016revisiting} and federated averaging approaches \cite{mcmahan2016communication}. We perform several different comparisons using the best hyperparameter selections for federated averaging and federated SGD. We compare client side computational costs when using deep models and demonstrate significantly lower computational burden on clients when training using our algorithm (see figure \ref{cifar100compute}). We also analyze the transmission cost of state-of-the-art deep networks including ResNet and VGG on CIFAR-10 and CIFAR-100. We demonstrate higher validation accuracy and faster convergence when considering a large number of clients. }

\nada{We demonstrate significant reductions in computation and communication bandwidth when comparing against federated SGD and federated averaging \cite{mcmahan2016communication}. Reduced computational requirements can be explained by the fact that while federated averaging requires forward pass and gradient computation for the entire neural network on the client, our method requires these computations for only the first few layers, significantly reducing the computational requirements (as shown in figure \ref{cifar100compute}). Even though federated averaging requires a lot fewer iterations than large-scale SGD, it is still outperformed by our method requiring only a fraction of computations on the client.}

\begin{figure*}[t]
\begin{subfigure} {0.5\textwidth}
\includegraphics[width=1\textwidth,trim={0.0cm 0 0.0cm 0},clip]{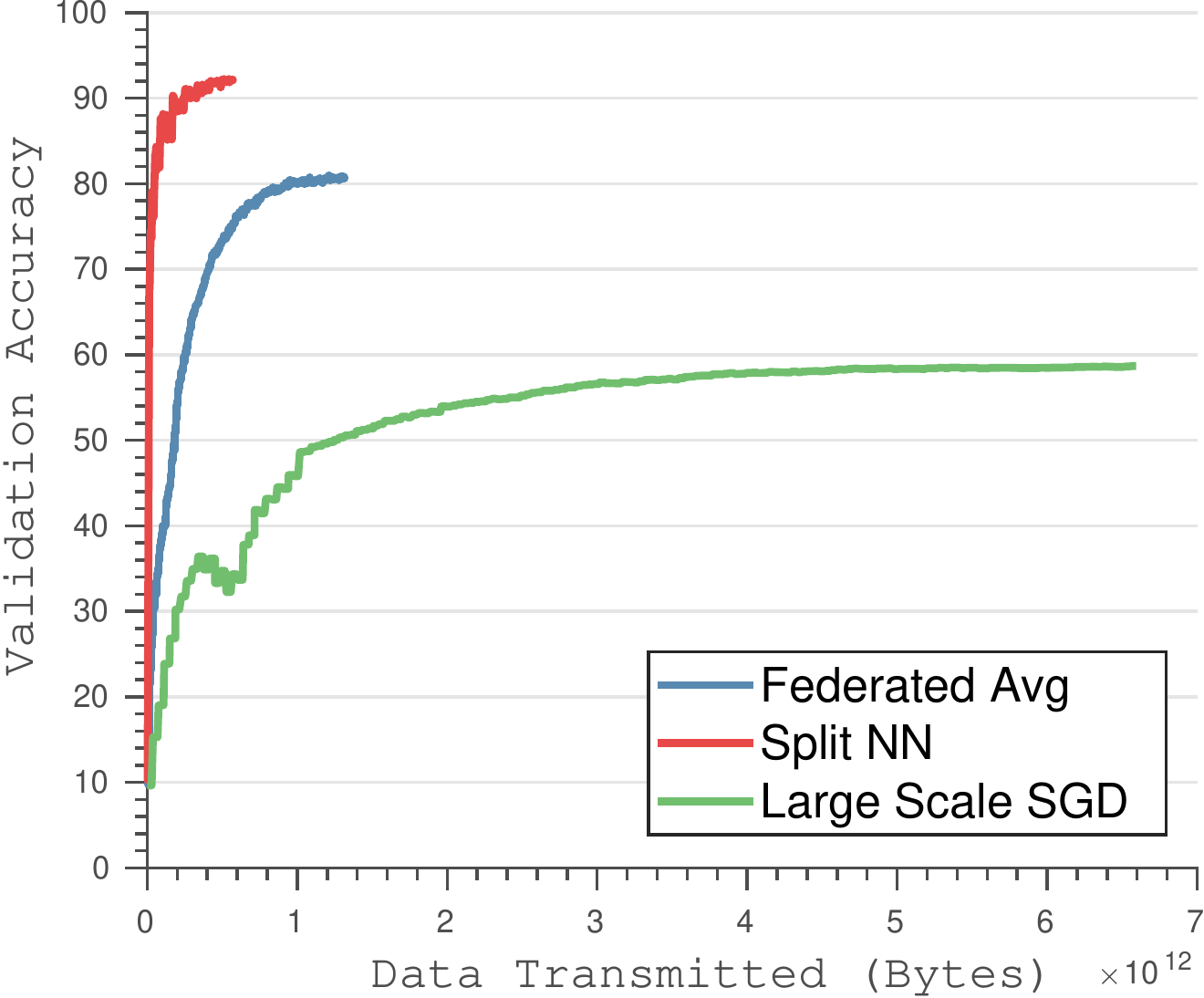}
\caption{Validation accuracy with transmitted data when training 500 clients using VGG over CIFAR-10.}
\label{cifar100communicationb}
\end{subfigure}
\begin{subfigure} {0.5\textwidth}
\includegraphics[width=1\textwidth,trim={0.0cm 0 0.0cm 0},clip]{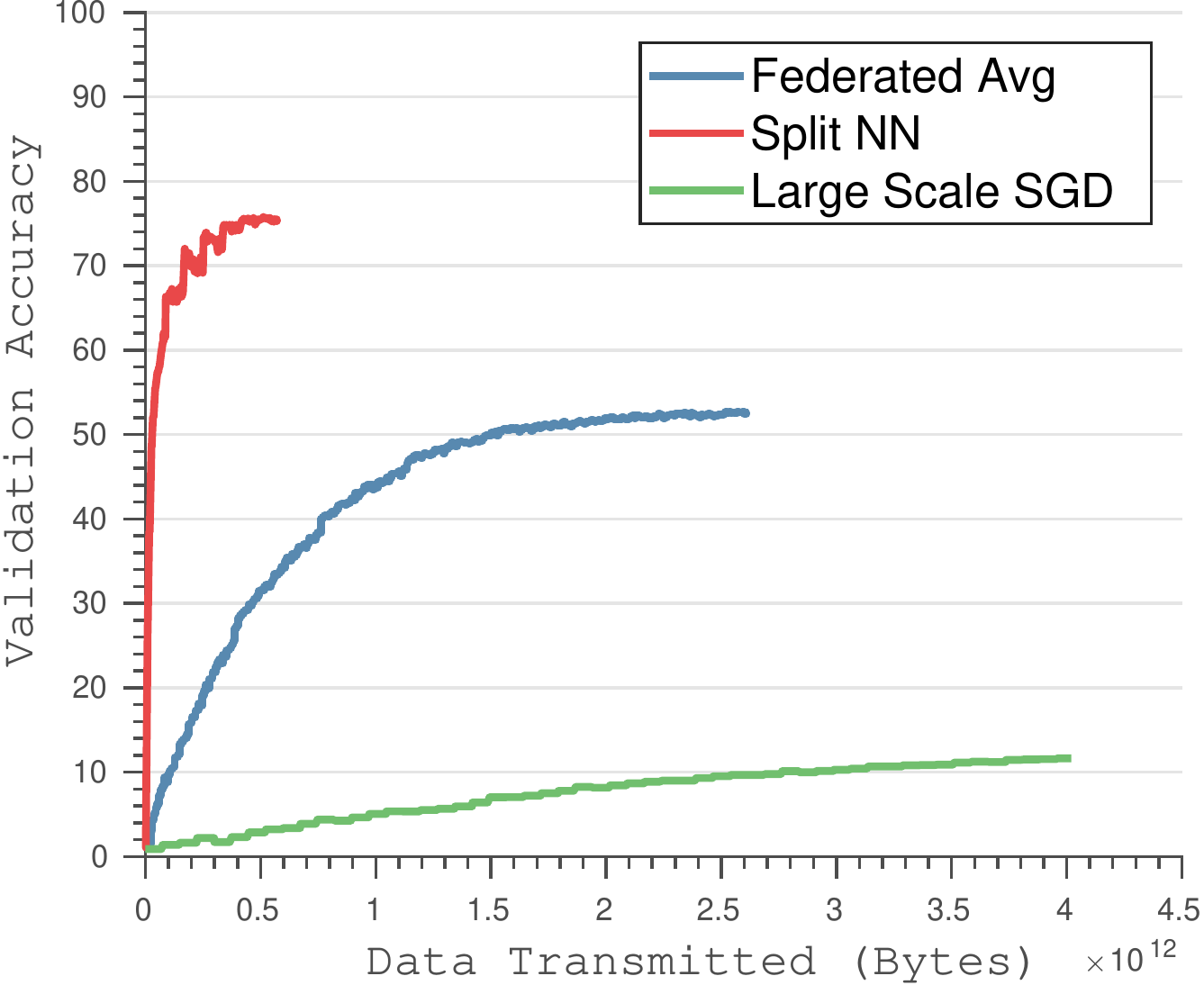}
\caption{Validation accuracy with transmitted data when training 500 clients using Resnet-50 over CIFAR-100.}
\label{cifarlimitation10client}
\end{subfigure}
\caption{Comparison of data transmission cost of our method against existing state of the art methods.}
\label{cifar100communication}
\end{figure*}

\nada{Reduction in communication bandwidth can be attributed to the fact that federated averaging involves transmitting the gradient updates for the entire neural network from all clients to a central server, accompanied by transmission of updated weights to every single client (please refer to figure \ref{cifar100communication}). While the federated averaging algorithm is able to converge in fewer transmission cycles, each transmission cycle requires huge amounts of data download and upload to the client and server. The split neural network algorithm reduces data transmitted by restricting the size of the client neural network to only the first few layers, thereby greatly reducing the total amount of data transmitted during training. Additionally, federated averaging fails to achieve optimal accuracy for higher numbers of clients since general non-convex optimization averaging models in parameter space could produce an arbitrarily bad model (phenomenon described in \cite{goodfellow2014qualitatively}).}

\subsection{Impact of amount of data on final accuracy}

\begin{table*}[t]
\centering
  \begin{tabular}{ l | p{4cm} | p{4cm} | p{4cm} }
    Dataset & Accuracy using 1 agent (10 \%) & Accuracy using 5 agents (50 \% of data) & Accuracy using all agents\\ \hline
    MNIST & 97.54 & 98.93 & 99.20\\
    CIFAR 10 & 72.53 & 89.05 & 92.45\\
    CIFAR 100 & 36.03 & 59.51 & 66.59\\
    ILSVRC 12 & 27.1 & 56.3* & 57.1\\
  \end{tabular}
\caption{Comparison on how accuracy improves as more data is added when training.}  
\label{accuracy_by_data}
\end{table*}

An important benefit of our method lies in its ability to combine multiple data-sources. When using deep neural networks, larger datasets have been shown to perform significantly better than smaller datasets. We experimentally demonstrate the benefits of pooling several agents by uniformly dividing dataset over 10 agents and training topologies using 1, 5 or 10 agents. We observe that adding more agents causes accuracy to improve significantly. Please see table \ref{accuracy_by_data} for analysis on how accuracy will improve as we add more data sources in real world scenarios.

\section{Conclusions and Future Work}
In this paper we present new methods to train deep neural networks over several data repositories. We also present algorithms on how to train neural networks without revealing actual raw data while reducing computational requirements on individual data sources. We describe how to modify this algorithm to work in semi-supervised modalities, greatly reducing number of labeled samples required for training. We provide mathematical guarantees for correctness of our algorithm. 

We devise a new protocol for easy implementation of our distributed training algorithm. We use popular computer vision datasets such as CIFAR-10 and ILSVRC12 for performance validation and show that our algorithm produces identical results to standard training procedures. We also show how this algorithm can be beneficial in low data scenarios by combining data from several resources. Such a method can be beneficial in training using proprietary data sources when data sharing is not possible. It can also be of value in areas such as biomedical imaging, when training deep neural network without revealing personal details of patients and minimizing the computation resources required on devices.

In this paper we describe a method to train a single network topology over several data repositories and a computational resource. A reasonable extension to this approach can be to train an ensemble of classifiers  by transmitting forward and backward tensors for all classifiers every iteration. A deep neural network classifier ensemble can comprise several individual deep neural network topologies which perform classification. The network topologies are trained individually by computing forward and backward functions for each neural network, and during the testing phase the results are combined using majority vote to produce classification. We can train such an ensemble by generating separate forward and backward propagation tensors for each neural network and transmitting them during each training iteration. This is equivalent to training individual networks one by one, but it saves time by combining iterations of various networks together. Ensemble classifiers have also been shown to be more secure against network copy attacks and have also been shown to perform better in real world applications \cite{granitto2005neural}.

In future work, a learned neural network could be shared using student-teacher methods for transferring information learned by neural network \cite{papernot2016semi}. After the training phases are over, Alice and Bob can use any publicly available dataset to train secondary (student) neural network using outputs from the primary (teacher) neural network. Alice can propagate the same training sample from the public dataset through the layers from the previously trained network and Bob can propagate them through its network. Bob can use the output of its layers to train the student network by doing forward-backward for the same data sample. This way, knowledge from the distributed trained network can be transferred to another network which can be shared for public use. Such algorithms can help in introducing deep learning in several areas such as health, products and finance where user data is an expensive commodity and needs to remain anonymized.

Tor like layer-by-layer computation could allow for training this network over multiple nodes with each node carrying only a few layers. Such a method could help protect not just the data but the identity of the person sharing the data and performing classification. In Tor like setup, additional entities \textit{Eve$_{0...M}$} are added which do not have access to data or complete network topology. Each Eve is provided with a few network layers $F_k^{eve} \gets {L_q,L_{q+1}...L_r}$. During forward propagation Alice computes $F_a$ and passes it to Eve$_0$, which then passes it to Eve$_1$ and so on until it reaches Eve$_M$. Eve$_M$ is analogous to the exit node in Tor network and it passes the tensor to Bob. Similarly, when backpropagating, Bob computes $loss$ and sends it to Eve$_M$, which sends it to Eve$_{M-1}$ and so on until it reaches Eve$_0$ and then Alice. The onion like organization of network layers can be used to keep the identity of Alice confidential.

We can also apply our algorithm on not just classification tasks but also on regression and segmentation tasks. We can also use this over LSTMs and Recurrent Neural Networks. Such neural networks can be easily tackled by using a different loss function (euclidean) on Bob's side when generating gradients.

\section*{References}
\bibliographystyle{elsarticle-num}

\end{document}